\documentclass{article}
\usepackage{ijcai21}

\usepackage{times}
\usepackage{soul}
\usepackage{url}
\usepackage[hidelinks]{hyperref}
\usepackage[utf8]{inputenc}
\usepackage[small]{caption}
\usepackage{graphicx}
\usepackage{amsmath}
\usepackage{amsthm}
\usepackage{booktabs}
\usepackage{multirow}
\usepackage{natbib}
\usepackage{dsfont}

\usepackage{amssymb}
\usepackage{calc}
\usepackage{newtxtext,newtxmath}

\usepackage{fancyhdr}

\usepackage{thmtools}

\declaretheorem{lemma}

\usepackage[table,dvipsnames]{xcolor}
\usepackage{epsfig}
\usepackage{wrapfig}
\usepackage{pgfplotstable}
\usepackage{pgfplots}
\usepgfplotslibrary{groupplots}
\pgfplotsset{compat=newest} %
\usepgfplotslibrary{groupplots}
\usepgfplotslibrary{dateplot}
\usepackage{mathrsfs,amsmath,mathtools}
\usepackage{bbm}
\usepackage{tikz}
\usetikzlibrary{decorations.text,calc,shapes,arrows,arrows.meta, positioning,shapes.misc,decorations.markings,decorations.markings,decorations.pathreplacing,matrix,spy}
\usepackage{rotating}

\usepackage{adjustbox}

\usepackage{subcaption}
\usepackage[noend]{algpseudocode}
\usepackage{algorithm}
\usepackage{siunitx}
\usepackage{nicefrac}
\usepackage{spverbatim}
\usepackage{seqsplit}

\usepackage{enumitem}
\setlist{nosep} %

\renewcommand{\cite}{\citep}

\makeatletter
\pgfplotsset{
    every axis x label/.append style={
        alias=current axis xlabel
    },
    legend pos/outer south/.style={
        /pgfplots/legend style={
            at={%
                (%
                \@ifundefined{pgf@sh@ns@current axis xlabel}%
                {xticklabel cs:0.5}%
                {current axis xlabel.south}%
                )%
            },
            anchor=north
        }
    }
}
\makeatother

\pgfplotscreateplotcyclelist{MyCyclelist}{%
  {Bittersweet, mark = *, dashed},
  {Blue, mark = square*, dashed},
  {Cyan, mark = diamond*, dashed},
  {DarkOrchid, mark = triangle*, dashed},
  {Green, mark = *, dashed},
  {Magenta, mark = square*, dashed},
  {Red, mark = diamond*, dashed},
  {Gray, mark = triangle*, dashed},
  {Black, mark = *, dashed},
    {Bittersweet, mark = square*, densely dotted},
  {Blue, mark = diamond*, densely dotted},
  {Cyan, mark = triangle*, densely dotted},
  {DarkOrchid, mark = *, densely dotted},
  {Green, mark = square*, densely dotted},
  {Magenta, mark = diamond*, densely dotted},
  {Red, mark = triangle*, densely dotted},
  {Gray, mark = *, densely dotted},
  {Black, mark = square*, densely dotted},
}

\algrenewcommand\algorithmicindent{0.75em}%

\everypar{\looseness=-1} %
\newcommand{\bigO}{\ensuremath\mathcal{O}}
\renewcommand{\bigO}{\ensuremath O}

\hypersetup{draft}

\title{Exact Acceleration of K-Means++ and K-Means$\|$}

\author{
    Edward Raff
    \affiliations
    Booz Allen Hamilton \\ 
    University of Maryland, Baltimore County
    \emails
    raff\_edward@bah.com, raff.edward@umbc.edu
}

\begin{document}

\maketitle

\begin{abstract}
K-Means++ and its distributed variant K-Means$\|$ have become de facto tools for selecting the initial seeds of K-means. While alternatives have been developed, the effectiveness, ease of implementation, and theoretical grounding of the K-means++ and $\|$ methods have made them difficult to ``best'' from a holistic perspective. By considering the limited opportunities within seed selection to perform pruning, we develop specialized triangle inequality pruning strategies and a dynamic priority queue to show the first acceleration of K-Means++ and K-Means$\|$ that is faster in run-time while being algorithmicly equivalent. For both algorithms we are able to reduce distance computations by over $500\times$. For K-means++ this results in up to a 17$\times$ speedup in run-time and a $551\times$ speedup for K-means$\|$. We achieve this with simple, but carefully chosen, modifications to known techniques which makes it easy to integrate our approach into existing implementations of these algorithms. 
\end{abstract}

\section{Introduction}

Before one can run the $K$-means algorithm, a prerequisite step is needed to select the initial $K$-\textit{seeds} to use as the initial estimate of the means. This seed selection step is critical to obtaining high quality results with the $K$-means algorithm. Selecting better initial centers $m_1, \ldots, m_K$  can improve the quality of the final  $K$-means clustering. A major step in developing better seed selection was the $K$-means++ algorithm. This was the first to show that the seeds it finds are log-optimal in expectation for solving the $K$-means problem~\cite{Arthur2007}. For a dataset with $n$ items $K$-means++ requires $\bigO(n K)$ distance computations. If $P$ processors are available $K$-means++ can be done in $\bigO(n K / P)$. However, the amount of communication overhead to do $K$-means in parallel is significant. To remedy this,~\citet{Bahmani:2012:SK:2180912.2180915} introduced $K$-means$\|$ which retains the $\bigO(n K / P)$ complexity and performs a constant factor more distance computations to significantly reduce the communication overhead while still yielding the same log-optimal results~\cite{pmlr-v70-bachem17b}. When working in a distributed environment, where communication must occur over the network, this can lead to large reductions in run-time~\cite{Bahmani:2012:SK:2180912.2180915}.

The cost of $K$-means++ has long been recognized as being an expensive but necessary step for better results~\cite{Hamerly2014}, with little progress on improvement. Modern accelerated versions of $K$-means clustering perform as few as $1.2$ total iterations of the dataset~\cite{Rysavy2016}, making $K$-means++ seed selection take up to 44\% of all distance computations. 
Outside of exact $K$-means clustering, faster seed selection can help improve stochastic variants of $K$-means~\cite{Bottou95convergenceproperties,Sculley2010} and is useful for applications like corset construction~\cite{pmlr-v37-bachem15}, change detection~\cite{Raff2020d}, tensor algorithms~\cite{10.5555/1813231.1813268}, clustering with Bergman divergences~\cite{10.1007/978-3-540-87481-2_11}, and Jensen divergences~\cite{Nielsen2015}. Applications with large $K$ have in particular been neglected, even though $K \geq 20,000$ is useful for scaling kernel methods~\cite{JMLR:v18:15-025}. 

In this work, we seek to accelerate the original $K$-means++ and $K$-means$\|$ algorithms so that we may obtain the same provably good results in less time without compromising on \textit{any} of the desirable qualities of $K$-means++ or $K$-means$\|$. We will review work related to our own in \autoref{sec:related_work}. Since the bottlenecks and approach to accelerating these two algorithms are different we will review their details and our approach to accelerating them sequentially. In respect to $K$-means++ in \autoref{sec:accel_kpp}, we show how simple application of the triangle inequality plus a novel dynamic priority queue allows us to avoid redundant computations and keep the cost of sampling new means low. In \autoref{sec:accel_kbb} we address $K$-means$\|$ and develop a new \textit{NearestInRange} query that allows us to successfully use a metric index to prune distance computations even though it is restricted to corpora normally too small to be useful with structures like KD-trees. We then perform empirical evaluation of our modifications in \autoref{sec:results} over a larger set of corpora with more diverse properties covering $K \in [32, 4096]$. In doing so, we observe that our accelerated algorithms succeed in requiring either the same or less time across all datasets and all values of $K$, making it a Pareto improvement. Finally, we will conclude in \autoref{sec:conclusion}. 

\section{Related Work} \label{sec:related_work}

Many prior works have looked at using the triangle inequality, $d(a,b) + d(b, c) \geq d(a, c)$, to accelerate the $K$-means algorithm. While the first work along this line was done by~\citet{Phillips2002}, it was first successfully popularized by~\citet{Elkan2003}. Since then, several works have attempted to build faster $K$-means clustering algorithms with better incorporation or tighter bounds developed through use of the triangle inequality~\cite{Hamerly2010,Ding:2015:YKD:3045118.3045181,Newling2016}. Despite the heavy use of the triangle inequality to accelerate $K$-means clustering, we are aware of no prior works that apply it to the seed selection step of $K$-means++ and $K$-means$\|$. We belive this is largely because these methods can not accelerate the first iteration of $K$-means, as they rely on the first iteration's result to accelerate subsequent iterations. Since $K$-means++ is effectively a single iteration of $K$-means, their approaches can not be directly applied to the seed selection step. 

In our work to accelerate $K$-means$\|$ using metric index structures a similar historical theme emerges. 
Prior works have looked at using index structures like KD-trees~\cite{Pelleg:1999:AEK:312129.312248} and Cover-trees~\cite{Curtin2017} to accelerate the $K$-means clustering algorithm, but did not look at the seed selection step. Similarly we will use a metric indices to accelerate $K$-means$\|$, but we will develop an enhanced nearest neighbor query that considers a maximum range 
to meaningfully prune even when using small values of $K$. 

Most work we are aware of focuses on extending or utilizing the $K$-means++ algorithm with few significant results on improving it. The most significant in this regard is the AFK-MC\cite{NIPS2016_6478} algorithm and its predecessor K-MC~\cite{Bachem:2016:AKS:3016100.3016103}.  Both can obtain initial seeds with the same quality as $K$-means++ with less distance computations but scale as $\bigO(n/P + m K^2)$, where $m$ is a budget factor. This makes them less effective when a large number of CPUs $P$ is available or when $K$ is large. Neither work factored in actual run-time.~\cite{NIPS2017_7104} showed that these implementations are actually 3.3$\times$ \textit{slower} when overheads are factored in. We consider run-time in our own work to show that our improvements materialize in practice.

\section{Accelerating $K$-Means++} \label{sec:accel_kpp}

We start with the $K$-means++ algorithm where we present detailed pseudo-code in \autoref{algo:kmeanspp}. We detail the method and how it works when each data point $x_i$ has with it an associated weight $w_i$, as this is required later on.  The algorithm begins by selecting an initial seed at random, and then assigning a new weight $\beta_i$ to each data point $x_i$, based on the squared distance of $x_i$ to the closest existing seed. At each iteration, we select a new seed to the set based on these weights and return once we have $k$ total seeds. This requires $k$ iterations through the dataset or size $n$ resulting in $\bigO(n \cdot k)$ distance computations. 
Note that we cache the distance between each point $x_i$ and it's closest mean into the variable $\alpha_i$. We will maintain this notation throughout the paper and use $\alpha_i$ as shorthand. 

\begin{algorithm}[!ht]
\caption{K-Means++}
\label{algo:kmeanspp}
\begin{algorithmic}[1]
\Require Desired number of  seeds $K$, data points $x_1, \ldots, x_n$, data weights $w_1, \ldots, w_n$
\State Weight of each data point $w_i \geq 0$
\State $\beta_i \gets w_i/\sum_{j=1}^n w_j, \forall i \in [1, n]$
\State $m_1 \gets x_i$ , where $i$ is selected with probability $\beta_i$
\State $k \gets 1$
\State $\alpha = \vec{\infty}$
\While{$k < K$}
    \For{$i \in [1, n]$}
        \State $\alpha_i \gets \min(\alpha_i, d(m_k, x_i))$
    \EndFor
    \State $Z \gets \sum_{i=1}^n w_i \cdot \alpha_i^2$
    \For{$i \in [1, n]$}
        \State $\beta_i \gets w_i \cdot \alpha_i^2/Z$
    \EndFor
    \State $k \gets k +1$
    \State $m_k \gets x_i$ , where $i$ is selected with probability $\beta_i$
\EndWhile
\State \Return  initial means $m_1, \ldots, m_K$
\end{algorithmic}
\end{algorithm}  

The first step toward improving the $K$-means++ algorithm is to filter out redundant distance computations. To do this, we note that at each iteration we compare the distance of each point $x_i$ to the newest mean $m_k$ against the previous closest mean $m_j$, where $1 \leq j < k$. That is, we need to determine if $d(x_i, m_k) < d(x_i, m_j)$. To do this, we can use \autoref{lema:bound} as introduced and proven by~\citet{Elkan2003}, 

\begin{lemma}\label{lema:bound}
 Let $x$ be a point and let $b$ and $c$ be centers. If
$d(b, c) \geq 2 d(x, b)$ then $d(x, c) \geq d(x, b) $
\end{lemma}

\subsection{Applying the Triangle Inequality}

We can use the distance between $m_p$ and $m_j$  to determine if computing $d(x_i, m_k)$ is a fruitless effort by checking if $d(m_j, m_k) >d(x_i, m_j)$. This is already available in the form of $\alpha_i$ as presented in \autoref{algo:kmeanspp}. We then only need to compute $d(m_j, m_k) \forall j < k$, of which there is intrinsically less than $k$ unique values at each iteration. Thus, we can compute $\gamma_j = d(m_j, m_k)$ once at the start of each loop, and we can re-use these $k$ values for all $n-k$ distance comparisons.

Applying this bound we can
avoid many redundant computations. As there are still $K$ total iterations to select $K$ means, each iteration will perform $k$ comparisons to previous means and $n-k$, we get at most $n$ distance comparisons per iteration making the worst case still $\bigO(n k)$ distance computations for the $K$-means++ algorithm.

\subsection{Avoiding Subnomral Slowdowns}

A non-trivial cost exists in lines 9-13 of \autoref{algo:kmeanspp} where we must compute the probability of selecting each point as the next mean and then perform the selection. This requires at least $3\cdot n$ floating point multiplications which can be a bottleneck in low dimensional problems.
This can be exasperated because squared distance to the closest center $\alpha_i^2$ naturally becomes very small as $k$ increases resulting in subnormalized floating point values. Subnormals (also called denormal) attempt to extend the precision of IEEE floats, but can cause $100\times$ slowdowns in computation~\cite{SubnormalOSIHPA06}. Depending on hardware, subnormals can also interfere with pipelining behavior and out-of-order execution, making a single subnormal computation highly detrimental to performance~\cite{Fog2016}. 
This is particularly problematic because pruning based on the triangle inequality works best on low dimensional problems, and the normalization step prevents us from realizing speedups in terms of total run-time. 

To circumvent this bottleneck, we develop a simple approach to create a \textit{dynamic} priority queue that allows us to sample the next mean accurately without having to interact with most of the samples per iteration. We start with the elegant sampling without replacement strategy introduced by~\citet{Efraimidis2006}. Given $n$ items $1, \ldots, n$ with weighted probabilities $w_1, \ldots, w_n$, it works by assigning each item $i$ a priority $\lambda_i w_i^{-1}$ where $\lambda_i$ is sampled from the Exponential distribution with $\lambda=1$ (i.e., $\lambda_i \sim \text{Exponential}(1)$). To select $K$ items without replacement, one selects the $K$ values with highest priority (smallest $\lambda_i w_i^{-1}$ values). This can normally be done with the quick-select algorithm in $\bigO(n)$ time. 

For $K$-means++ seeding we want to instead use the priority $\lambda_i w_i^{-1} \alpha_i^{-2}$ in order to produce random samples. The term $ w_i^{-1} \alpha_i^{-2}$ acts as the weight for datum $i$ being selected, and it is a combination of the original relative weight of the datum $w_i$ and the squared distance to the nearest seed $\alpha_i^2$. At the start we sample $\lambda_i \sim \text{Exponential}(1)$ once. During each round, we update all $\alpha_i$ values and leave $\lambda_i$ fixed. It is trivial to see that this does not alter the expectation of any point being selected conditioned on the point $i$ already being removed. This is because all $\lambda_i$ are sampled independently, and so the removal of any $\lambda_i$ does not impact the relative weights of any other point. Thus, we can use the weighted sampling without replacement strategy of ~\citet{Efraimidis2006} to select the seeds. We performed a sanity check by implementing this naive approach and making no other changes. This resulted in the same quality solutions over many trials with the same statistical mean and variance. 

At first glance, this strategy obtains no benefit as the value of $\alpha_i$ will change on each iteration. Each value of $\alpha_i$ changing means that the relative ordering of all remaining priorities $\lambda_i w_i^{-1} \alpha_i^{-2}$ will also change. This requires a full quick-select run on each iteration to discover the new maximum priority item. However, we note that $\alpha_i$ can only decrease with each iteration, and thus the priority of any given sample either remains constant or decreases. Our first contribution is the realization that this property can be exploited to reduce the cost of sampling so that only a subset of priorities need to be considered to sample the next point. 

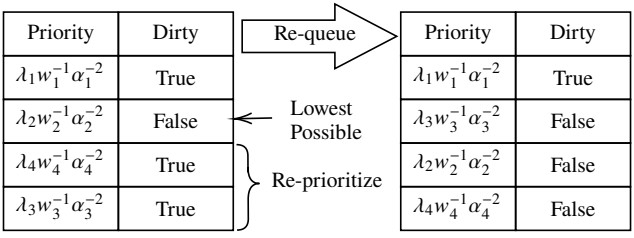
\begin{figure}
    \centering
    \adjustbox{max width=\columnwidth}{%
        \tikzset{every picture/.style={line width=0.75pt}} %

\begin{tikzpicture}[x=0.75pt,y=0.75pt,yscale=-1,xscale=1]

\draw   (12.5,123) -- (83.5,123) -- (83.5,150) -- (12.5,150) -- cycle ;
\draw   (83.5,123) -- (154.5,123) -- (154.5,150) -- (83.5,150) -- cycle ;
\draw   (12.5,96) -- (83.5,96) -- (83.5,123) -- (12.5,123) -- cycle ;
\draw   (83.5,96) -- (154.5,96) -- (154.5,123) -- (83.5,123) -- cycle ;

\draw   (12.5,69) -- (83.5,69) -- (83.5,96) -- (12.5,96) -- cycle ;
\draw   (83.5,69) -- (154.5,69) -- (154.5,96) -- (83.5,96) -- cycle ;
\draw   (12.5,42) -- (83.5,42) -- (83.5,69) -- (12.5,69) -- cycle ;
\draw   (83.5,42) -- (154.5,42) -- (154.5,69) -- (83.5,69) -- cycle ;

\draw   (12.5,15) -- (83.5,15) -- (83.5,42) -- (12.5,42) -- cycle ;
\draw   (83.5,15) -- (154.5,15) -- (154.5,42) -- (83.5,42) -- cycle ;
\draw   (156.83,149.83) .. controls (161.5,149.8) and (163.82,147.46) .. (163.79,142.79) -- (163.72,131.06) .. controls (163.67,124.39) and (165.98,121.04) .. (170.65,121.01) .. controls (165.98,121.04) and (163.63,117.73) .. (163.59,111.06)(163.61,114.06) -- (163.55,103.96) .. controls (163.52,99.29) and (161.17,96.97) .. (156.5,97) ;
\draw    (178.5,80.5) -- (157,80.96) ;
\draw [shift={(155,81)}, rotate = 358.78] [color={rgb, 255:red, 0; green, 0; blue, 0 }  ][line width=0.75]    (10.93,-3.29) .. controls (6.95,-1.4) and (3.31,-0.3) .. (0,0) .. controls (3.31,0.3) and (6.95,1.4) .. (10.93,3.29)   ;

\draw   (258,123) -- (329,123) -- (329,150) -- (258,150) -- cycle ;
\draw   (329,123) -- (400,123) -- (400,150) -- (329,150) -- cycle ;
\draw   (258,96) -- (329,96) -- (329,123) -- (258,123) -- cycle ;
\draw   (329,96) -- (400,96) -- (400,123) -- (329,123) -- cycle ;

\draw   (258,69) -- (329,69) -- (329,96) -- (258,96) -- cycle ;
\draw   (329,69) -- (400,69) -- (400,96) -- (329,96) -- cycle ;
\draw   (258,42) -- (329,42) -- (329,69) -- (258,69) -- cycle ;
\draw   (329,42) -- (400,42) -- (400,69) -- (329,69) -- cycle ;

\draw   (258,15) -- (329,15) -- (329,42) -- (258,42) -- cycle ;
\draw   (329,15) -- (400,15) -- (400,42) -- (329,42) -- cycle ;
\draw   (160,20) -- (217.3,20) -- (217.3,10) -- (255.5,30) -- (217.3,50) -- (217.3,40) -- (160,40) -- cycle ;

\draw (47.5,135) node    {$\lambda _{3} w^{-1}_{3} \alpha ^{-2}_{3}$};
\draw (119,136.5) node   [align=left] {True};
\draw (47.5,108) node    {$\lambda _{4} w^{-1}_{4} \alpha ^{-2}_{4}$};
\draw (119,109.5) node   [align=left] {True};
\draw (47.5,81) node    {$\lambda _{2} w^{-1}_{2} \alpha ^{-2}_{2}$};
\draw (119,82.5) node   [align=left] {False};
\draw (119,55.5) node   [align=left] {True};
\draw (47.5,54) node    {$\lambda _{1} w^{-1}_{1} \alpha ^{-2}_{1}$};
\draw (119,28.5) node   [align=left] {Dirty};
\draw (48,28.5) node   [align=left] {Priority};
\draw (212.5,120) node   [align=left] {Re-prioritize};
\draw (212,81.5) node   [align=left] {Lowest \\Possible};
\draw (293,135) node    {$\lambda _{4} w^{-1}_{4} \alpha ^{-2}_{4}$};
\draw (364.5,136.5) node   [align=left] {False};
\draw (293,81) node    {$\lambda _{3} w^{-1}_{3} \alpha ^{-2}_{3}$};
\draw (364.5,82.5) node   [align=left] {False};
\draw (364.5,28.5) node   [align=left] {Dirty};
\draw (293.5,28.5) node   [align=left] {Priority};
\draw (364.5,55.5) node   [align=left] {True};
\draw (293,54) node    {$\lambda _{1} w^{-1}_{1} \alpha ^{-2}_{1}$};
\draw (364.5,109.5) node   [align=left] {False};
\draw (293,108) node    {$\lambda _{2} w^{-1}_{2} \alpha ^{-2}_{2}$};
\draw (205,29.5) node   [align=left] {Re-queue};

\end{tikzpicture}
    }
    \caption{Example of priority re-queueing strategy for $n=4$ items. Initially, it is not clear if items 2, 3, or 4 are the next to sample. All dirty items are removed from the queue until we reach a clean item and then re-inserted after fixing their priorities. We do not need to consider any item after the first clean item.
    }
    \label{fig:requeue}
\end{figure}

We can instead create a priority queue using a standard binary heap to select the next smallest value of $\lambda_i w_i^{-1} \alpha_i^{-2}$ and maintain a marker if the priority of an item $i$ has become \textit{dirty}. An item is dirty if and only if the item has a higher priority than it actually should. If there is a clean item $z$ in the queue, then all items with a lower apparent priority than $z$ must have a true priority that is still lower than $z$. Thus, we need only fix the priority of items higher than $z$. 

See \autoref{fig:requeue} for an example of this queue for a dataset of $n=4$ items. Item $2$ is clean, and all items with a higher priority (3 and 4) are dirty. That means item 2 has the lowest possible priority that \textit{could} be the next true sample because it is possible the values of items $3$ and $4$ will become larger (read, lower priority) once the updated values of $\alpha_3$ and $\alpha_4$ are computed. Thus, we can remove all items in the queue until we reach item $2$ and then re-insert them into the queue with their correct priorities. In this hypothetical example, item $4$ still had a lower priority after updating, and so will become the next mean when we them remove it from the queue. Item 1 occurred after item 2 because it had a lower priority. Even though item 1 was dirty, we did not need to consider it because its priority can only decrease once $\alpha_1$ is updated. Because Item 2 was clean, its priority will not change, and there is no possibility of item 1 being selected. 

\subsection{Accelerated $K$-Means++}

\begin{algorithm}[!ht]
\caption{Our Accelerated K-Means++}
\label{algo:kmeanspp_accel}
\begin{algorithmic}[1]
\Require Desired number of  seeds $K$, data points $x_1, \ldots, x_n$, data weights $w_1, \ldots, w_n$
\State $\lambda_i \sim \text{Exponential}(1), \forall i \in [1, n]$
\State Weight of each data point $w_i \geq 0$
\State Priority Queue $Q$ with each index $i$ given priority $\lambda_i/w_i$
\State $\text{dirty}_i \gets $ False
\State $m_1 \gets x_{Q.\Call{Pop}{ }}$ 
\State $\alpha = \vec{\infty}$, $k \gets 1$,  $\phi_i \gets 0$
\For{$k \in [1, K)$} \Comment{\textcolor{blue}{For each new center $k$}}
    \For{$j \in [1, k)$} \Comment{\textcolor{blue}{Get distance to previous centers}}
        \State $\gamma_j \gets d(m_k, m_j)$
    \EndFor
    \For{$i \in [1, n]$}
        \If{$\frac{1}{2}\gamma_{\phi_i} \geq \alpha_i$}
            \State \textbf{continue} \Comment{\textcolor{blue}{Pruned by \autoref{lema:bound}}}
        \EndIf
        \If{$d(m_k, x_i) < \alpha_i$}
            \State $\alpha_i \gets  d(m_k, x_i)$
            \State $\phi_i \gets k$
            \State $\text{dirty}_i \gets $ True \Comment{\textcolor{blue}{Priority may now be too high}}
        \EndIf
    \EndFor
    \State Create new stack $S$
    \While{$\text{dirty}_{Q.\Call{Peek}{ }}$} \Comment{\textcolor{blue}{All items that \textit{could} be selected}}
        \State $i \gets Q.\Call{Pop}{ }$
        \State $S.\Call{Push}{i}$
    \EndWhile
    \For{$i \in S$} \Comment{\textcolor{blue}{Update true priority}}
        \State $Q.\Call{Push}{i, \lambda_i / \left(w_i \cdot \alpha_i^2 \right) }$
        \State $\text{dirty}_i \gets $ False
    \EndFor
    \State $m_k \gets x_{Q.\Call{Pop}{ }}$ \Comment{\textcolor{blue}{Select new mean by clean top priority}}
\EndFor
\State \Return initial means $m_1, \ldots, m_K$
\end{algorithmic}
\end{algorithm}  

The final algorithm that performs the accelerated computation is given in \autoref{algo:kmeanspp_accel}. Lines 8-12 take care to avoid redundant distance computations, and lines 16-23 ensure that the dynamic priority queue allows us to select the next mean without considering all $n-k$ remaining candidates. Combined, we are able to regularly gain reductions both in terms of total time taken as well as the number of distance computations required. Through the use of our dynamic priority queue we find that we regularly consider less than 1\% of total remaining $n-k$ items. This is important when we work with low-dimension datasets. When the dimension is very small (e.g., $d=2$ for longitude/latitude data is a common use case), there is little computational cost in the distance computations themselves, and so much of the bottleneck in runtime is contained within the sampling process. Our dynamic queue avoids this bottleneck allowing us to realize the benefits of reduced distance computations.

\section{Accelerating $K$-Means$\|$} \label{sec:accel_kbb}

Now we turn our attention to the $K$-means$\|$ algorithm detailed in \autoref{algo:kmeansbb}. While $K$-means$\|$ requires more distance computations, it is preferred in distributed environments because it requires less communication which is a significant bottleneck for $K$-means++~\cite{Bahmani:2012:SK:2180912.2180915}. It works by reducing the $K$ rounds of communication to a fixed number of $R \ll K$ rounds, yet still obtains the log-optimal results of $K$-means++~\cite{pmlr-v70-bachem17b}. In each of the rounds, $\ell$ new means are sampled based on the weighted un-normalized probability $\ell w_i \alpha_i^2$. With the standard defaults of $R=5$ and $\ell=2 K$, we end up with an expected $R 2 K > K$ total means. These $R \cdot \ell$ potential means are weighted by the number of points that they are closest to and then passed to the $K$-means++ algorithm to reduce them to a final set of $K$ means, which produces the final result. Note this last step requires $\bigO(K^2)$ distance computations when naively using \autoref{algo:kmeanspp}, making it necessary to accelerate the $K$-means++ algorithm in order to effectively accelerate $K$-means$\|$ for datasets with large $K$. 

\begin{algorithm}[!ht]
\caption{K-Means$\|$ }
\label{algo:kmeansbb}
\begin{algorithmic}[1]
\Require  Desired number of  seeds $K$, $x_1, \ldots, x_n$, data weights $w_1, \ldots, w_n$, rounds $R$, oversampling factor $\ell$
\State Weight of each data point $w_i \geq 0$
\State $\beta_i \gets w_i/\sum_{j=1}^n w_j, \forall i \in [1, n]$
\State $c_1 \gets x_i$ , where $i$ is selected with probability $\beta_i$
\State $k \gets 1$, $k_{\mathit{prev}}\gets 0$, $\alpha = \vec{\infty}$
\For{$r \in [1, R]$}
    \For{$i \in [1, n]$}
        \For{$j \in (k_{\mathit{prev}}, k]$}
            \State $\alpha_i \gets \min\left(\alpha_i, d\left(c_j, x_i\right)\right)$
        \EndFor
    \EndFor
    \State $k_{\mathit{prev}}\gets k$
    \State $Z \gets \sum_{i=1}^n w_i \cdot \alpha_i^2$
    \For{$i \in [1, n]$}
        \If{$p \sim \text{Ber}(\min(1, \ell \cdot w_i \cdot \alpha_i^2 / z))$ is true}
            \State $k \gets k+1$,  $c_k \gets x_i$, $\alpha_i \gets 0$
        \EndIf
    \EndFor
\EndFor
\State Let $w_i' \gets \sum_{j=1}^n w_j \cdot \mathds{1}[d(c_i, x_j) = \alpha_j]$ \Comment{\textcolor{blue}{Weight set to number of points closest to center $c_i$}}
\State \Return  K-Means++($K$, $c_1, \ldots, c_k$, $w_1', \ldots, w_k'$) \Comment\textcolor{blue}{{Run \autoref{algo:kmeanspp}}}
\end{algorithmic}
\end{algorithm}  

Since $ K < R \cdot \ell \ll n$, the final step of running $K$-means++ is not overbearing to run on a single compute node, and the sampling procedure is no longer a bottleneck that requires subversion. In a distributed setting, the $\ell$ new means selected are broadcast out to all worker nodes, which is possible because $\ell \ll n$, and thus requires limited communication overhead. However, the ability to use the triangle inequality becomes less obvious. Using the same approach as before, similar to~\citet{Elkan2003}, would require $\bigO(K^2)$ pairwise distances computations between the new and old means, and more book-keeping overhead that would reduce the effectiveness of avoiding distance computations. 

Another strategy uses an algorithm like the Cover-Tree that accelerates nearest neighbor searches and supports the removal of data points from the index~\cite{Beygelzimer2006,Izbicki2015}. Then, we could perform an all-points nearest neighbor search~\cite{Curtin2013}. However, we are unaware of any approach that has produced a distributed cover-tree algorithm that would not run into the same communication overheads that prevents the standard $K$-means++ from working in this scenario. As such, it does not appear to be a worth while strategy.

\subsection{Nearest In Range Queries}

Another approach would be to fit an index structure $\mathscr{C}$ to only the $\ell$ new points, and for each non-mean $x_i$ find its nearest potentially new assignment by querying $\mathscr{C}$. Since $\ell$ is $\bigO(K)$ this is too small a dataset for pruning to be effective with current methods.

To remedy this, we 
note that we have additional information available to perform the search. The value $\alpha_i$ which indicates the distance of point $x_i$ to its closest current mean. As such, we introduce a \textit{NearestInRange} search that returns the nearest neighbor to a query point $q$ against an index $\mathscr{C}$ if it is within a radius of $r$ to the query. Since most points $x_i$ will not change ownership in a given iteration, a NearestInRange search could be able to prune out the entire search tree, and it will increase its effectiveness even if $K$ is small. 

To do this, we use the Vantage Point tree (VP) algorithm~\cite{Yianilos1993} because it is fast to construct, has low overhead which makes it competitive with other algorithms such as KD-trees and Cover-trees~\cite{Raff2018_metric_index}, and simple to augment with our new NearestInRange search. The pseudo-code for the standard VP search is given in \autoref{algo:nearest_in_range}, where \textit{GetChild}, \textit{Search}, and \textit{Best} are auxiliary functions used by the \textit{Nearest} function to implement a standard nearest neighbor search. The VP has a left and right child, and it uses a value $\tau$ to keep track of the distance to the nearest neighbor found. It also maintains two pairs of bounds, $\mathit{near}_{\mathit{low}}, \mathit{near}_{\mathit{high}}$ indicating the shortest and farthest distance to the points in the left child and $\mathit{far}_{\mathit{low}}, \mathit{far}_{high}$ 
do the same for the right child.

\begin{algorithm}[!ht]
\caption{Nearest Neighbor Search in VP Tree}
\label{algo:nearest_in_range}
\begin{algorithmic}[1]
\Function{GetChild}{ $low$}
    \If{$low = true$}
        \State \Return left child
    \EndIf
    \State \Return right child \Comment{\textcolor{blue}{Else, return other}}
\EndFunction
\Function{Search}{$r$, $\tau$, $low$}
    \If{$low = true$}
        \State $a \gets \mathit{near}_{\mathit{low}}, b \gets \mathit{near}_{\mathit{high}}$
    \Else
        \State  $a \gets \mathit{far}_{\mathit{low}}, b \gets \mathit{far}_{\mathit{high}}$
    \EndIf
    \State \Return $a - \tau < r < b + \tau$ \Comment{\textcolor{blue}{i.e., is this True or False?}}
\EndFunction
\Function{Best}{$\tau$, $\tau'$, $ID$, $ID'$}
    \If{$\tau <  \tau'$}
        \State \Return $\tau, ID$
    \EndIf
    \State \Return $\tau', ID'$ \Comment{\textcolor{blue}{Else, return other}}
\EndFunction
\Function{Nearest}{$q$, $\tau$, $ID$}
  \State $r \gets d(p, q)$
  \State $\tau, ID \gets \Call{Best}{\tau, r, ID, p}$
  \State $m \gets \frac{\mathit{near}_{\mathit{high}} + \mathit{far}_{\mathit{low}}}{2}$
  \State $lf \gets r < m$ \Comment{\textcolor{blue}{True/False, search near/left child first?}}
  \If{\Call{Search}{$r, \tau, lf$}}
    \State $\tau', ID' \gets $\Call{GetChild}{ $lf$}.\Call{Nearest}{$q, \tau, ID$}
    \State $\tau, ID \gets \Call{Best}{\tau, \tau', ID, ID'}$
  \EndIf
  \If{\Call{Search}{$r, \tau, \neg{}lf$}}
    \State $\tau', ID' \gets $\Call{GetChild}{ $\neg{}lf$}.\Call{Nearest}{$q, \tau, ID$}
    \State $\tau, ID \gets \Call{Best}{\tau, \tau', ID, ID'}$
  \EndIf
  \State \Return $\tau, ID$
\EndFunction
\Function{NearestInRange}{$q$, $maxRange$} 
    \State \Return \Call{Nearest}{$q, maxRange, -1$} \Comment{\textcolor{blue}{This simple function, used in-place of Nearest, is our contribution. }}
\EndFunction
\end{algorithmic}
\end{algorithm}  

A standard Nearest Neighbor search calls the Nearest function with $\tau = \infty$, and the bound is updated as the search progresses when it fails to prune a branch. Our contribution is simple. The \textit{NearestInRange} function instead sets $\tau = r$, the minimum viable radius. It is easy to verify that this can only monotonically improve the pruning rate of each search. Since $\tau$ bounds the distance to the nearest neighbor, and we know from the $\boldsymbol{\alpha}$ values an upper-bound on the distance to the nearest neighbor, the modification remains correct. The rest of the algorithm remains unaltered, and can simply terminate the search faster due to a meaningful initial bound.

\begin{algorithm}[!ht]
\caption{Our Accelerated K-Means$\|$}
\label{algo:kmeansbb_accel}
\begin{algorithmic}[1]
\Require  Desired number of  seeds $K$, $x_1, \ldots, x_n$, data weights $w_1, \ldots, w_n$, rounds $R$, oversampling factor $\ell$
\State Weight of each data point $w_i \geq 0$
\State $\beta_i \gets w_i/\sum_{j=1}^n w_j, \forall i \in [1, n]$
\State $c_1 \gets x_i$ , where $i$ is selected with probability $\beta_i$
\State $\alpha = \vec{\infty}$, $k_{\mathit{prev}} \gets 0$, $k \gets 1$
\For{$r \in [1, R]$}
    \State $\mathscr{C} \gets $ new index built from $\{c_{k_{\mathit{prev}}}, \ldots, c_k\}$
    \For{$i \in [1, n]$}
        \State $j \gets $ $\mathscr{C}.$\Call{NearestInRange}{$x_i, \alpha_i$}
        \If{$j \geq 0$} %
            \State $\alpha_i \gets d(c_j, x_i)$
        \EndIf
    \EndFor
    \State $k_{\mathit{prev}} \gets k$
    \State $Z \gets \sum_{i=1}^n w_i \cdot \alpha_i^2$
    \For{$i \in [1, n]$}
        \If{$p \sim \text{Ber}(\min(1, \ell \cdot w_i \cdot \alpha_i^2 / Z))$ is true}
        \State $k \gets k+1$,  $c_k \gets x_i$, $\alpha_i \gets 0$
        \EndIf
    \EndFor
\EndFor
\State Let $w_i' \gets \sum_{j=1}^n w_j \cdot \mathds{1}[d(c_i, x_j) = \alpha_j]$ \Comment{\textcolor{blue}{Weight set to number of points closest to center $c_i$}}
\State \Return  K-Means++($K$, $c_1, \ldots, c_k$, $w_1', \ldots, w_k'$) \Comment{\textcolor{blue}{Run \autoref{algo:kmeanspp_accel}}}
\end{algorithmic}
\end{algorithm}  

Thus, to build an accelerated $K$-means$\|$ we build an index  $\mathscr{C}$ on the newly selected means. We do comparisons against that filtered with our \textit{NearestInRange} search, as detailed in \autoref{algo:kmeansbb_accel}. For the first iteration, the loop on lines 7-11 will be fast with only $c_1$ to determine the initial distribution, and on every subsequent round we have a meaningful value of $\alpha_i$ that can be used to accelerate the search. If none of the $\ell$ new candidates $c_{k_{\mathit{prev}}}, \ldots, c_k$ are within a distance of $\alpha_i$ to each point $x_i$, then the \textit{NearestInRange} function will return a negative index which can be skipped.

In addition, we use our accelerated $K$-means++ algorithm \autoref{algo:kmeanspp_accel} in the final step rather than the standard algorithm. This allows us to accelerate all parts of the $K$-means$\|$ method while also keeping the simplicity and low communication cost of the original design. The Vantage Point tree is a small index since it is built upon a small dataset of $\ell$ points, and the index can be sent to every worker node in a cluster in the exact same manner.

\section{Experimental Results} \label{sec:results}

Now that we have detailed the methods by which we accelerate the $K$-means++ and $K$-means$\|$ algorithms, we will evaluate their effectiveness. The two measures we are concerned with are the following: 1) reducing the total number of distance computations and 2) the total run-time spent. Measuring distance computations gives us an upper-bound on potential effectiveness of our algorithm, and allows us to compare approaches in an implementation and hardware independent manner. Measuring the run-time gives us information about the ultimate goal, which is to reduce the time it takes to obtain $K$ seeds. However, it is sensitive to the hardware in use, the language the approach is implemented in, and the relative skills of program authors. For this work we used 
the JSAT library~\cite{JMLR:v18:16-131}. 
The $K$-means++ algorithm was provided by this framework, and we implemented the $K$-means$\|$ and accelerated versions of both algorithms using JSAT. This way all comparisons with respect to run-time and the $K$-means++ and $\|$ algorithms presented are directly comparable. Our implementations have been contributed to the JSAT library for public use. 

Prior works that have investigated alternatives to $K$-means++ have generally explored only a few datasets with $D<100$ features and less than 4 values of $K$, sometimes testing only one value of $K$ per dataset~\cite{Bachem:2016:AKS:3016100.3016103}. For example, while MNIST is regularly tested in seed selection, it is usually projected down to 50 dimensions first~\cite{Hamerly2010} due to being difficult to accelerate. 

Since our goal is to produce accelerated versions of these algorithms that are uniformly better, we attempt to test over a wide selection of reasonable scenarios. In \autoref{tbl:datasets} we show the 11 datasets we use, with $D \in [3, 780]$, and $n$ covering four orders of magnitude. We will test $K \in [32, 4096]$ covering each power of two so that we may understand the behavior as $K$ changes and to make sure we produce an improvement even when $K$ is small. To the best of our knowledge, this is a larger number of datasets, range and values of $K$, and range and values of $D$ to be tested compared to prior work\footnote{We are aware of no prior work in this space that has considered $D > 1024$, where pruning methods are unlikely to succeed due to the curse of dimensionality. We consider this reasonable and beyond scope, as such scenarios are usually sparse and best handled by topic models like LDA.}.

\begin{wraptable}[17]{r}{0.55\columnwidth}
\centering
\vspace{-10pt}
\adjustbox{width=0.55\columnwidth}{%
\begin{tabular}{@{}lrr@{}}
\toprule
\multicolumn{1}{c}{Dataset} & \multicolumn{1}{c}{$n$} & \multicolumn{1}{c}{$D$} \\ \midrule
Phishing       & 11055                 & 68                    \\
cod-rna      & 59535                 & 8                     \\
MNIST      & 60000                 & 780                   \\
aloi        & 108000                & 128                   \\
Range-Queries    & 200000                & 8                     \\
Skin/NoSkin  & 245057                & 3                     \\
covertype    & 581012                & 54                    \\
SUSY        & 5000000               & 18                    \\
Activity Rec.  & 33741500              & 5                     \\
HIGGS      & 11000000              & 28                    \\
Web\footnote{\url{http://webscope.sandbox.yahoo.com/}, R6A - Yahoo! Front Page Today Module User Click Log Dataset}                         & 45811883              & 5                     \\ 
\bottomrule
\end{tabular}
}\caption{Datasets used. Left is the dataset, ordered by number of samples ($n$). Right most column indicates the number of features $D$. } \label{tbl:datasets}
\end{wraptable}

Unless stated otherwise, all experiments were done with a single CPU core from an iMac with a 3.5 GHz Intel i5 CPU with 64 GB of RAM. 
The phishing dataset is only tested up to $K=2048$, because at $K=4096$ we would be selecting over 1/4 of the dataset as means, at which point the purpose of $K$-means++ style seeding is being defeated by selecting too large a portion of the corpus. All results are averaged over 5 runs, and took four months to complete in our compute environment. 

\subsection{$K$-Means++ Results}

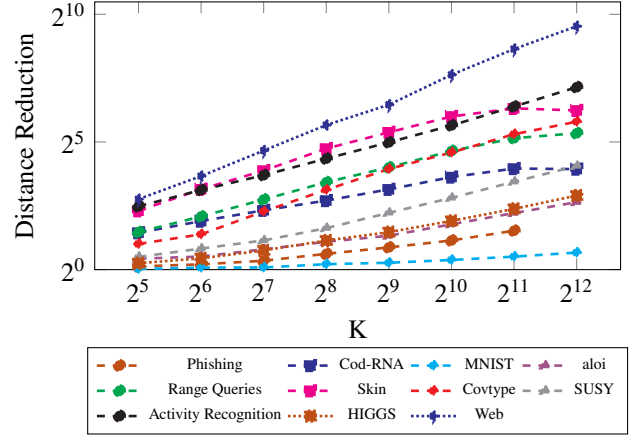
\begin{figure}[!htb]
\begin{center}
\centering
\begin{tikzpicture}[]
\begin{axis}[
    xlabel=K,
    ylabel=Distance Reduction,
    legend style={font=\tiny},
    legend pos=outer south,
    legend columns=4,
    xmode=log,
    log basis x={2},
    ymode=log,
    log basis y={2},
    ymin=1.0,
    height=0.6\columnwidth,
    width=\columnwidth,
    cycle list name=MyCyclelist
    ]
    
    \addplot+[line width=1.0pt] table [x=k, y expr=\thisrow{dist_avg}/\thisrow{tia_dist_avg}, col sep=comma] {results/phishing_kpp_11055_68.csv};
    \addlegendentry{Phishing}

    \addplot+[line width=1.0pt] table [x=k, y expr=\thisrow{dist_avg}/\thisrow{tia_dist_avg}, col sep=comma] {results/cod-rna_kpp_59535_8.csv};
    \addlegendentry{Cod-RNA}

    \addplot+[line width=1.0pt] table [x=k, y expr=\thisrow{dist_avg}/\thisrow{tia_dist_avg}, col sep=comma] {results/mnist_kpp_60000_780.csv};
    \addlegendentry{MNIST}
    
    \addplot+[line width=1.0pt] table [x=k, y expr=\thisrow{dist_avg}/\thisrow{tia_dist_avg}, col sep=comma] {results/aloi.scale_kpp_108000_128.csv};
    \addlegendentry{aloi}

    \addplot+[line width=1.0pt] table [x=k, y expr=\thisrow{dist_avg}/\thisrow{tia_dist_avg}, col sep=comma] {results/Range-Queries-Aggregates.csv_kpp_200000_8.csv};
    \addlegendentry{Range Queries}

    \addplot+[line width=1.0pt] table [x=k, y expr=\thisrow{dist_avg}/\thisrow{tia_dist_avg}, col sep=comma] {results/skin_nonskin_kpp_245057_3.csv};
    \addlegendentry{Skin}

    \addplot+[line width=1.0pt] table [x=k, y expr=\thisrow{dist_avg}/\thisrow{tia_dist_avg}, col sep=comma] {results/covtype.libsvm.binary_kpp_581012_54.csv};
    \addlegendentry{Covtype}

    \addplot+[line width=1.0pt] table [x=k, y expr=\thisrow{dist_avg}/\thisrow{tia_dist_avg}, col sep=comma] {results/SUSY_kpp_5000000_18.csv};
    \addlegendentry{SUSY}
    
    \addplot+[line width=1.0pt] table [x=k, y expr=\thisrow{dist_avg}/\thisrow{tia_dist_avg}, col sep=comma] {results/hetero_activity_recognition.csv_kpp_33741500_5.csv};
    \addlegendentry{Activity Recognition}

    \addplot+[line width=1.0pt] table [x=k, y expr=\thisrow{dist_avg}/\thisrow{tia_dist_avg}, col sep=comma] {results/HIGGS_kpp_11000000_28.csv};
    \addlegendentry{HIGGS}
    
    \addplot+[line width=1.0pt] table [x=k, y expr=\thisrow{dist_avg}/\thisrow{tia_dist_avg}, col sep=comma] {results/web48m.csv_kpp_45811882_5.csv};
    \addlegendentry{Web}

\end{axis}

\end{tikzpicture}
\end{center}
\caption{Factor reduction in distance computations for our accelerated $K$-means++ algorithm  compared to original. 
}
\label{fig:dist_reduction_kpp}
\end{figure}

We start with the $K$-means++ results with the reduction in distance computations shown in \autoref{fig:dist_reduction_kpp}. In the worst case for $K=32$ on the MNIST dataset, we still have to do 98\% of the distance computations as the standard algorithm, but this improved to only 63\% by $K=4096$. The best case is observed with the Web dataset, starting out with only 15\% of the distance computations at $K=32$ and only 0.1\% by $K=4096$, a 739$\times$ improvement. 

Across all the datasets, we see that the factor reduction in distance computations is a monotonic improvement for $K$-means++. We never see any case where our accelerated approach performs more distance computations than the naive approach. This confirms our decision to do an extra $k-1$ distance computation between the newest mean $m_k$ and the previous means $m_1, \ldots, m_{k-1}$.

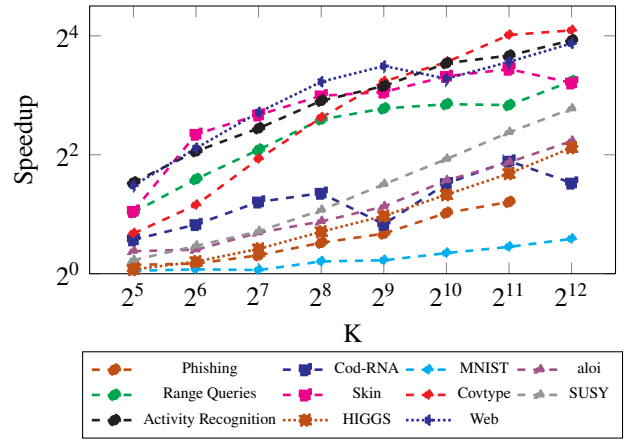
\begin{figure}[!htb]
\begin{center}
\centering
\begin{tikzpicture}[]
\begin{axis}[
    xlabel=K,
    ylabel=Speedup,
    legend style={font=\tiny},
    legend pos=outer south,
    legend columns=4,
    xmode=log,
    log basis x={2},
    ymode=log,
    log basis y={2},
    ymin=1.0,
    height=0.6\columnwidth,
    width=\columnwidth,
    cycle list name=MyCyclelist
    ]
    
    \addplot+[line width=1.0pt] table [x=k, y expr=\thisrow{time_avg}/\thisrow{tia_time_avg}, col sep=comma] {results/phishing_kpp_11055_68.csv};
    \addlegendentry{Phishing}

    \addplot+[line width=1.0pt] table [x=k, y expr=\thisrow{time_avg}/\thisrow{tia_time_avg}, col sep=comma] {results/cod-rna_kpp_59535_8.csv};
    \addlegendentry{Cod-RNA}

    \addplot+[line width=1.0pt] table [x=k, y expr=\thisrow{time_avg}/\thisrow{tia_time_avg}, col sep=comma] {results/mnist_kpp_60000_780.csv};
    \addlegendentry{MNIST}
    
    \addplot+[line width=1.0pt] table [x=k, y expr=\thisrow{time_avg}/\thisrow{tia_time_avg}, col sep=comma] {results/aloi.scale_kpp_108000_128.csv};
    \addlegendentry{aloi}

    \addplot+[line width=1.0pt] table [x=k, y expr=\thisrow{time_avg}/\thisrow{tia_time_avg}, col sep=comma] {results/Range-Queries-Aggregates.csv_kpp_200000_8.csv};
    \addlegendentry{Range Queries}

    \addplot+[line width=1.0pt] table [x=k, y expr=\thisrow{time_avg}/\thisrow{tia_time_avg}, col sep=comma] {results/skin_nonskin_kpp_245057_3.csv};
    \addlegendentry{Skin}

    \addplot+[line width=1.0pt] table [x=k, y expr=\thisrow{time_avg}/\thisrow{tia_time_avg}, col sep=comma] {results/covtype.libsvm.binary_kpp_581012_54.csv};
    \addlegendentry{Covtype}

    \addplot+[line width=1.0pt] table [x=k, y expr=\thisrow{time_avg}/\thisrow{tia_time_avg}, col sep=comma] {results/SUSY_kpp_5000000_18.csv};
    \addlegendentry{SUSY}
    
    \addplot+[line width=1.0pt] table [x=k, y expr=\thisrow{time_avg}/\thisrow{tia_time_avg}, col sep=comma] {results/hetero_activity_recognition.csv_kpp_33741500_5.csv};
    \addlegendentry{Activity Recognition}

    \addplot+[line width=1.0pt] table [x=k, y expr=\thisrow{time_avg}/\thisrow{tia_time_avg}, col sep=comma] {results/HIGGS_kpp_11000000_28.csv};
    \addlegendentry{HIGGS}
    
    \addplot+[line width=1.0pt] table [x=k, y expr=\thisrow{time_avg}/\thisrow{tia_time_avg}, col sep=comma] {results/web48m.csv_kpp_45811882_5.csv};
    \addlegendentry{Web}

\end{axis}

\end{tikzpicture}
\end{center}
\caption{Run-time Speedup for our accelerated  $K$-means++ algorithm compared to the standard algorithm. 
}
\label{fig:time_reduction_kpp}
\end{figure}

As we noted in the design of our accelerated variant, we must avoid over-emphasising the performance of just reduced distance computations as the cost of re-normalizing the distribution to sample the next mean is a non-trivial cost. This is especially true when we are able to reduce the distance computations by $\geq 16\times$ for several of our datasets. The results showing the run-time reduction are presented in \autoref{fig:time_reduction_kpp}. 

In all cases, our accelerated version of $K$-means++ is always faster than the standard algorithm. As expected, MNIST has the lowest speedup based on the number of distance computations avoided. At $K=32$ we achieved only a 3.4\% reduction in time but was $1.5\times$ faster by $K=4096$. 

\subsubsection*{Dynamic Priority Impact}

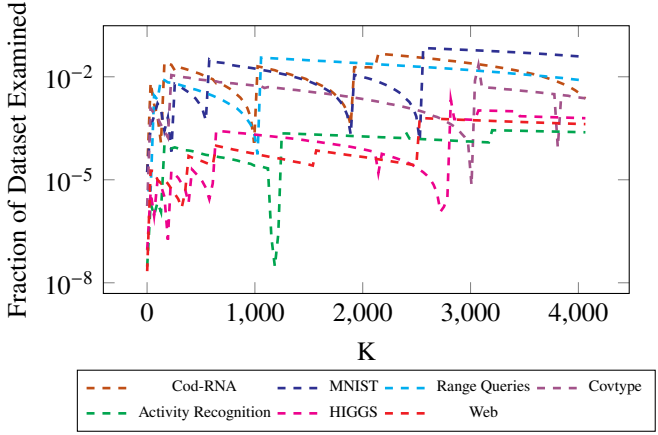
\begin{figure}[!h]
\begin{center}
\centering
\begin{tikzpicture}[]
\begin{axis}[
    xlabel=K,
    ylabel=Fraction of Dataset Examined,
    legend style={font=\tiny},
    legend pos=outer south,
    legend columns=4,
    ymode=log,
    log basis y={10},
    height=0.6\columnwidth,
    width=\columnwidth,
    cycle list name=MyCyclelist
    ]

    \addplot+[line width=1.0pt,each nth point={32},mark=none] table [x=k, y expr=\thisrow{codRNA}+1.0/59535, col sep=comma] {results/requeue.csv};
    \addlegendentry{Cod-RNA}

    \addplot+[line width=1.0pt,each nth point={32},mark=none] table [x=k, y expr=\thisrow{mnist}+1.0/60000, col sep=comma] {results/requeue.csv};
    \addlegendentry{MNIST}

    \addplot+[line width=1.0pt,each nth point={32},mark=none] table [x=k, y expr=\thisrow{rangeQueries}+1.0/200000, col sep=comma] {results/requeue.csv};
    \addlegendentry{Range Queries}

    \addplot+[line width=1.0pt,each nth point={32},mark=none] table [x=k, y expr=\thisrow{covtype}+1.0/581012, col sep=comma] {results/requeue.csv};
    \addlegendentry{Covtype}

    \addplot+[line width=1.0pt,each nth point={32},mark=none] table [x=k, y expr=\thisrow{activityRecognition}+1.0/33741500, col sep=comma] {results/requeue.csv};
    \addlegendentry{Activity Recognition}

    \addplot+[line width=1.0pt,each nth point={32},mark=none] table [x=k, y expr=\thisrow{higgs}+1.0/11000000, col sep=comma] {results/requeue.csv};
    \addlegendentry{HIGGS}
    
    \addplot+[line width=1.0pt,each nth point={32},mark=none] table [x=k, y expr=\thisrow{web}+1.0/45811883, col sep=comma] {results/requeue.csv};
    \addlegendentry{Web}

\end{axis}

\end{tikzpicture}
\end{center}
\caption{The fraction of the remaining candidates that need to be examined (y-axis, log scale) to select the $k$'th mean (x-axis, linear scale) using our dynamic priority queue. 
}
\label{fig:req_effectiveness}
\end{figure}

Since the normalization step is non-trivial, especially when $D$ is small, we see that the actual speedup in run-time is not as strongly correlated with the dimension $D$. The Covertype dataset ($D=54$) had the 4th largest reduction in distance computations, but it had the largest reduction in run-time with a $17\times$ improvement at $K=4096$. Our ability to still obtain real speedups on these datasets is because our dynamic priority queue  allows us to consider only a small subset of the dataset to accurately select the next weighted random mean. This can be seen in \autoref{fig:req_effectiveness}, where a subset of the datasets are shown with the fraction of the corpus examined on the y-axis. As the datasets get larger our dynamic queue generally becomes more effective, thus reducing the number of points that need to be checked to $\leq 1\%$.

To confirm that our dynamic priority queue's results are meaningful, we perform an ablation of \autoref{algo:kmeanspp_accel} where the dynamic priority queue on lines 18-23 are replaced with the standard sampling code from \autoref{algo:kmeanspp}. We run both versions and record the speedup when our dynamic queue is used in \autoref{tbl:dynamic_queue_speedup} for $K=4096$. Here we can see that with the exception of the cod-rna dataset, where there is a $<2\%$ slowdown (on the fastest dataset to run), our approach gives a $5\%-231\%$ speedup in all other cases with a median improvement of 20\%. 
\begin{wraptable}[21]{r}{0.55\columnwidth}
\centering
\caption{Ablation testing of speedup from using our new dynamic priority queue to perform seed selection at every iteration. Positive values indicate faster results using our dynamic queue, where our pruning from \autoref{algo:kmeanspp_accel} was used with/without the dynamic queue.} \label{tbl:dynamic_queue_speedup}
\centering
\adjustbox{max width=0.55\columnwidth}{%
\begin{tabular}{@{}lc@{}}
\toprule
\multicolumn{1}{c}{Dataset} & Speedup \\ \midrule
cod-rna                     & 0.983   \\
Phishing                    & 2.313   \\
MNIST                       & 1.059   \\
aloi                        & 1.059   \\
Range-Queries               & 1.342   \\
Skin/NoSkin                 & 1.217   \\
covtype                     & 1.877   \\
SUSY                        & 1.259   \\
Activity Recognition        & 1.207   \\
HIGGS                       & 1.279   \\
Web                         & 1.725   \\ \bottomrule
\end{tabular}
}
\end{wraptable}
We also note that for all $K <4096$ we still observe benefits to our queue, but the variance does increase to the degree of speedup. We did not observe any performance regressions larger than 3\% in extended testing.

\subsection{$K$-Means$\|$ Results}

In \autoref{fig:dist_reduction_kbb} we show the factor reduction in distance computations, which mirrors the overal trends of \autoref{fig:dist_reduction_kpp}. 
The results have improved by 
an additional 
$\approx2-4\times$ with a 579$\times$ reduction in distance computations on the Activity Recognition dataset. 
The MNIST dataset still had the least improvement, but still obtained a more significant 88\% reduction in distance computations at $K=32$. 

\begin{figure}[!htb]
\begin{center}
\centering
\begin{tikzpicture}[]
\begin{axis}[
    xlabel=K,
    ylabel=Distance Reduction,
    legend style={font=\tiny},
    legend pos=outer south,
    legend columns=4,
    xmode=log,
    log basis x={2},
    ymode=log,
    log basis y={2},
    ymin=0.0,
    height=0.6\columnwidth,
    width=\columnwidth,
    cycle list name=MyCyclelist
    ]
    
    \addplot+[line width=1.0pt] table [x=k, y expr=\thisrow{dist_avg}/\thisrow{tia_dist_avg}, col sep=comma] {results/phishing_kbb_11055_68.csv};
    \addlegendentry{Phishing}

    \addplot+[line width=1.0pt] table [x=k, y expr=\thisrow{dist_avg}/\thisrow{tia_dist_avg}, col sep=comma] {results/cod-rna_kbb_59535_8.csv};
    \addlegendentry{Cod-RNA}

    \addplot+[line width=1.0pt] table [x=k, y expr=\thisrow{dist_avg}/\thisrow{tia_dist_avg}, col sep=comma] {results/mnist_kbb_60000_780.csv};
    \addlegendentry{MNIST}
    
    \addplot+[line width=1.0pt] table [x=k, y expr=\thisrow{dist_avg}/\thisrow{tia_dist_avg}, col sep=comma] {results/aloi.scale_kbb_108000_128.csv};
    \addlegendentry{aloi}

    \addplot+[line width=1.0pt] table [x=k, y expr=\thisrow{dist_avg}/\thisrow{tia_dist_avg}, col sep=comma] {results/Range-Queries-Aggregates.csv_kbb_200000_8.csv};
    \addlegendentry{Range Queries}

    \addplot+[line width=1.0pt] table [x=k, y expr=\thisrow{dist_avg}/\thisrow{tia_dist_avg}, col sep=comma] {results/skin_nonskin_kbb_245057_3.csv};
    \addlegendentry{Skin}

    \addplot+[line width=1.0pt] table [x=k, y expr=\thisrow{dist_avg}/\thisrow{tia_dist_avg}, col sep=comma] {results/covtype.libsvm.binary_kbb_581012_54.csv};
    \addlegendentry{Covtype}

    \addplot+[line width=1.0pt] table [x=k, y expr=\thisrow{dist_avg}/\thisrow{tia_dist_avg}, col sep=comma] {results/SUSY_kbb_5000000_18.csv};
    \addlegendentry{SUSY}
    
    \addplot+[line width=1.0pt] table [x=k, y expr=\thisrow{dist_avg}/\thisrow{tia_dist_avg}, col sep=comma] {results/hetero_activity_recognition.csv_kbb_33741500_5.csv};
    \addlegendentry{Activity Recognition}

    \addplot+[line width=1.0pt] table [x=k, y expr=\thisrow{dist_avg}/\thisrow{tia_dist_avg}, col sep=comma] {results/HIGGS_kbb_11000000_28.csv};
    \addlegendentry{HIGGS}
    
    \addplot+[line width=1.0pt] table [x=k, y expr=\thisrow{dist_avg}/\thisrow{tia_dist_avg}, col sep=comma] {results/web48m.csv_kbb_45811882_5.csv};
    \addlegendentry{Web}

\end{axis}

\end{tikzpicture}
\end{center}
\caption{Factor Reduction in Distance Computations needed to perform $K$-means$\|$ seed selection. Larger is better. 
}
\label{fig:dist_reduction_kbb}
\end{figure}
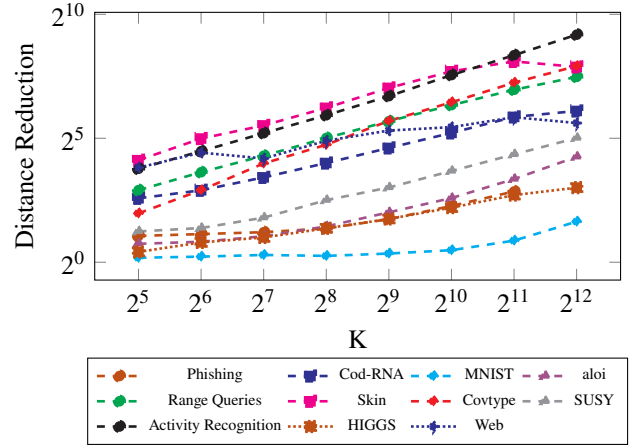

\begin{figure}[!htb]
\begin{center}
\centering
\begin{tikzpicture}[]
\begin{axis}[
    xlabel=K,
    ylabel=Speedup,
    legend style={font=\tiny},
    legend pos=outer south,
    legend columns=4,
    xmode=log,
    log basis x={2},
    ymode=log,
    log basis y={2},
    ymin=0.0,
    height=0.6\columnwidth,
    width=\columnwidth,
    cycle list name=MyCyclelist
    ]
    
    \addplot+[line width=1.0pt] table [x=k, y expr=\thisrow{time_avg}/\thisrow{tia_time_avg}, col sep=comma] {results/phishing_kbb_11055_68.csv};
    \addlegendentry{Phishing}

    \addplot+[line width=1.0pt] table [x=k, y expr=\thisrow{time_avg}/\thisrow{tia_time_avg}, col sep=comma] {results/cod-rna_kbb_59535_8.csv};
    \addlegendentry{Cod-RNA}

    \addplot+[line width=1.0pt] table [x=k, y expr=\thisrow{time_avg}/\thisrow{tia_time_avg}, col sep=comma] {results/mnist_kbb_60000_780.csv};
    \addlegendentry{MNIST}
    
    \addplot+[line width=1.0pt] table [x=k, y expr=\thisrow{time_avg}/\thisrow{tia_time_avg}, col sep=comma] {results/aloi.scale_kbb_108000_128.csv};
    \addlegendentry{aloi}

    \addplot+[line width=1.0pt] table [x=k, y expr=\thisrow{time_avg}/\thisrow{tia_time_avg}, col sep=comma] {results/Range-Queries-Aggregates.csv_kbb_200000_8.csv};
    \addlegendentry{Range Queries}

    \addplot+[line width=1.0pt] table [x=k, y expr=\thisrow{time_avg}/\thisrow{tia_time_avg}, col sep=comma] {results/skin_nonskin_kbb_245057_3.csv};
    \addlegendentry{Skin}

    \addplot+[line width=1.0pt] table [x=k, y expr=\thisrow{time_avg}/\thisrow{tia_time_avg}, col sep=comma] {results/covtype.libsvm.binary_kbb_581012_54.csv};
    \addlegendentry{Covtype}

    \addplot+[line width=1.0pt] table [x=k, y expr=\thisrow{time_avg}/\thisrow{tia_time_avg}, col sep=comma] {results/SUSY_kbb_5000000_18.csv};
    \addlegendentry{SUSY}
    
    \addplot+[line width=1.0pt] table [x=k, y expr=\thisrow{time_avg}/\thisrow{tia_time_avg}, col sep=comma] {results/hetero_activity_recognition.csv_kbb_33741500_5.csv};
    \addlegendentry{Activity Recognition}

    \addplot+[line width=1.0pt] table [x=k, y expr=\thisrow{time_avg}/\thisrow{tia_time_avg}, col sep=comma] {results/HIGGS_kbb_11000000_28.csv};
    \addlegendentry{HIGGS}
    
    \addplot+[line width=1.0pt] table [x=k, y expr=\thisrow{time_avg}/\thisrow{tia_time_avg}, col sep=comma] {results/web48m.csv_kbb_45811882_5.csv};
    \addlegendentry{Web}

\end{axis}

\end{tikzpicture}
\end{center}
\caption{Run-time speedup for our accelerated $K$-means$\|$ algorithm compared to the standard algorithm. Larger is better. 
}
\label{fig:speedup_kbb}
\end{figure}
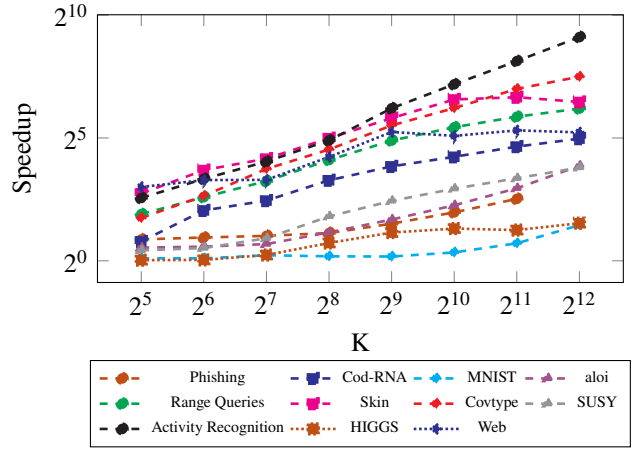

The approximately $4\times$ improvement in distance computations also carries over to the total run-time, as shown in \autoref{fig:speedup_kbb}. We observe a more consistent behavior because the cost of normalizing and sampling the new means is reduced to only $R=5$ rounds of sampling. 
Where our accelerated $K$-means++ had the relative improvement drop significantly for small $D<10$ datasets due to this overhead, our accelerated $K$-means$\|$ algorithm sees the ordering remain relatively stable. For example, the Activity Recognition dataset enjoys the greatest reduction in distance computations as well as run-time, and the 579$\times$ reduction in distance computations closely matches the 551$\times$ reduction in run-time.  The HIGGS dataset has the lowest improvement in run-time with a $1.02\times$ speedup at $K=32$ and 2.9$\times$ at $K=4096$. We also note that the NearestInRange query provided an additional $1.5 - 4\times$ speedup in most cases, but was highly dependent on the dataset and value of $K$.

\section{Conclusion} \label{sec:conclusion}

Leveraging simple modifications and a novel priority queue, we show the first method that delivers equal or better run-time in theory (less distance computations) and practice (less run-time). None of our changes impact the function of $K$-means++ or $K$-means$\|$, allowing us to retain existing properties.

\section*{Acknowledgements}

I would like to thank Ashley Klein, Drew Farris, and Frank Ferraro for reviewing early drafts and their valuable feedback.

\bibliographystyle{named}
\bibliography{short}

\end{document}